\documentclass[10pt,twocolumn,letterpaper]{article}

\usepackage[final]{cvpr}      %
\usepackage{xcolor}
\usepackage{bbding}
\usepackage{multicol}
\usepackage{multirow}
\usepackage{xcolor}

\definecolor{cvprblue}{rgb}{0.21,0.49,0.74}
\usepackage[pagebackref,breaklinks,colorlinks,allcolors=cvprblue]{hyperref}

\def\paperID{*****} %
\def\confName{****}
\def\confYear{****}

\newcommand\dataset{OW-COD}
\newcommand\method{MR-GDINO}

\title{{\method}: Efficient Open-World Continual Object Detection}
\author{
   Bowen Dong\textsuperscript{\rm 1,2}\quad Zitong Huang\textsuperscript{\rm 1}\quad Guanglei Yang\textsuperscript{\rm 1}\quad Lei Zhang\textsuperscript{\rm 2}\quad  Wangmeng Zuo\textsuperscript{\rm 1} \\
   \textsuperscript{\rm 1}Harbin Institute of Technology \quad \textsuperscript{\rm 2}The Hong Kong Polytechnic University\\
   \small{\{cndongsky, zitonghuang99\}@gmail.com \quad cslzhang@comp.polyu.edu.hk \quad wmzuo@hit.edu.cn}
}

\begin{document}
\twocolumn[{%
\renewcommand\twocolumn[1][]{#1}%
\maketitle
\vspace{-2.8em}
\begin{center}
   \captionsetup{type=figure}
   \includegraphics[width=0.99\textwidth]{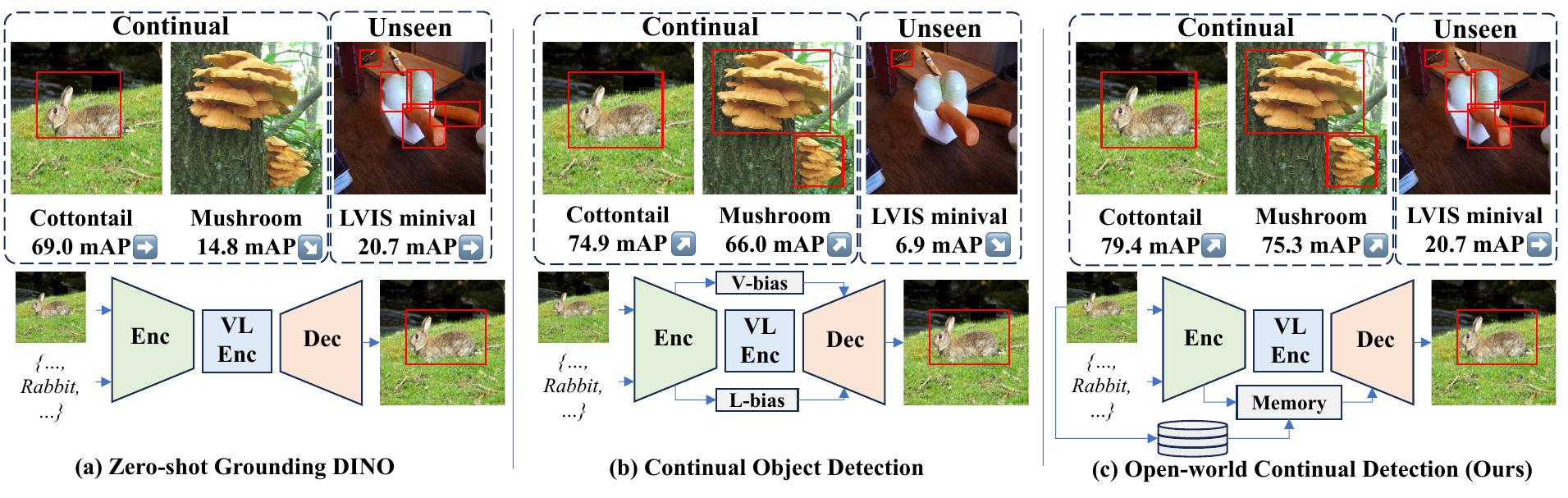}%
   \vspace{-1.2em}
   \captionof{figure}{
   (a) Pretrained open-world (OW) detectors~\cite{liu2023grounding} show strong generalization abilities on unseen data but cannot benefit from few-shot annotations. (b) Continual detectors~\cite{deng2024zero} built on OW detectors with continual learning show improved mAP on seen data but suffer from  forgetting for unseen objects. (c) Our OW continual detector {\method} via memory and retrieval improves detection abilities on seen classes while preserving OW abilities on unseen classes.
   }\label{fig:intro}
\end{center}%
\vspace{-0.5em}
}]

\begin{abstract}
Open-world (OW) recognition and detection models show strong zero- and few-shot adaptation abilities, inspiring their use as initializations in continual learning methods to improve performance. Despite promising results on seen classes, such OW abilities on unseen classes are largely degenerated due to catastrophic forgetting. To tackle this challenge, we propose an open-world continual object detection task, requiring detectors to generalize to old, new, and unseen categories in continual learning scenarios. Based on this task, we present a challenging yet practical {\dataset}  benchmark to assess detection abilities. The goal is to motivate OW detectors to simultaneously {\textbf{preserve} learned classes, \textbf{adapt} to new classes, and \textbf{maintain} open-world capabilities} under few-shot adaptations. To mitigate forgetting in unseen categories, we propose {\method}, a strong, efficient and scalable baseline via \textbf{memory and retrieval} mechanisms within a highly scalable memory pool. Experimental results show that existing continual detectors suffer from severe forgetting for both seen and unseen categories. In contrast, {\method} largely mitigates forgetting with only 0.1\% activated extra parameters, achieving state-of-the-art performance for old, new, and unseen categories. Code is available at \url{https://m1saka.moe/owcod/}.

\end{abstract}
    
\section{Introduction}\label{sec:intro}
Open-world (OW) recognition and detection models~\cite{radford2021learning,yao2022filip,zhai2023sigmoid,wang2024ov,zhang2022glipv2,bansal2018zero} has shown remarkable progress in effectively recognize and localize generalized objects~\cite{wang2023v3det,gupta2019lvis,lin2014microsoft,zhai2019large} with different granularity~\cite{li2023semantic,wang2023hierarchical}. 
By learning with vast semantic-rich data~\cite{objects365,wang2023v3det,lin2014microsoft,kazemzadeh2014referitgame,yu2016modeling,kamath2021mdetr,changpinyo2021cc12m,zhang2024controlvideo}, even without supervised by any bounding boxes from unseen classes, OW models (\emph{e.g.}, object detection networks) can generalize well under a \emph{zero-shot learning paradigm in open-world scenarios} (\emph{e.g.}, zero-shot Grounding DINO for OW in Fig~\ref{fig:intro}(a)). 
Benefited from highly generalized feature representation of detectors, such detectors can be also adapt to new classes via many- or few-shot fine-tuning~\cite{li2021grounded,liu2023grounding,adapter-few-shot-det,hu2022lora,he2022towards}, thus performing better on target classes. 

By sequentially repeating the fine-tuning procedure, one can formulate the updated OW detectors as a continual learning paradigm~\cite{wang2022learning,zhou2024class}. 
This formulation is intuitive and more practical than traditional open-world learning~\cite{li2024open} or continual learning~\cite{zhou2024class} by two-fold.
First, pretrained open-world (OW) detectors offer a robust initial representation that supports both zero-shot learning and rapid adaptation. 
Second, OW detectors are designed to encounter various out-of-distribution scenarios, but still suffer from performance drops by shifts in domain~\cite{chen2023sam,tang2023can,MedSAM} and previously unseen categories~\cite{wang2021wanderlust,pmlr-v162-guo22g,continual_autodrive,huang2024class}. 
Those models with fast adaptation can effectively address this issue, enhancing performance under real-world deployment conditions. 
Thus, we anticipate that OW detectors will preserve the advantages of open-world learning and demonstrate strong generalization across both known and novel categories.

As shown in Fig~\ref{fig:intro}(b),
prior studies~\cite{deng2024zero,liu2023grounding,li2024open,li2021grounded} on OW detectors~\cite{liu2023grounding} in continual learning
make feature representations strictly adapt to target classes domain~\cite{dong2023lpt,he2022towards}, and break original well-aligned visual-language representation. 
Though obtaining better performance than zero-shot OW detectors among seen categories, traditional continual learning frameworks for OW detectors still suffer degree of catastrophic forgetting for previously seen categories. Moreover, after continual adaptations on seen categories, the unseen categories detection capabilities of obtained detectors degenerate.
This limitation constrains the applicability of OW detectors in real-world scenarios.
To tackle this challenge, our research pursues two objectives, including: 1) assessing catastrophic forgetting in OW detectors across various learning frameworks, and 2) developing continual learning strategies specifically tailored for OW detectors for promising detection abilities on seen and unseen categories. 

To this end, we propose the \emph{open-world continual object detection task}, which requires optimized open-world detectors to simultaneously \textbf{preserve} knowledge of old classes, \textbf{adapt} to new classes, and \textbf{maintain} detection capabilities for unseen classes under continual few-shot adaptations.
Due to the lack of proper evaluation toolkits, building upon this task, we propose a challenging yet practical benchmark {\dataset}, specifically designed to evaluate anti-catastrophic forgetting capabilities across old, new, and unseen categories within continual learning frameworks for OW detectors.
Specifically, {\dataset} includes two groups of data. The former is few-shot training data with corresponding evaluation samples from various domains~\cite{li2021grounded}, which are sequentially utilized to optimize OW detectors via continual learning paradigm, and evaluate the detection performance for both old and new seen classes under class-incremental settings. 
And the latter is large-scale open-world object detection evaluation data~\cite{gupta2019lvis}, which are used to assess the detection accuracy of unseen categories against catastrophic forgetting. 
The combination of evaluation for both seen and unseen categories fits the goal of our task, and provides a comprehensive benchmark for continual learning frameworks under open-world detection scenarios.

Based on {\dataset} benchmark, we construct a strong baseline method to achieve the goal of our task.
Based on prior studies~\cite{li2021grounded,dou2022coarse}, we argue that explicit visual-language interaction module is the key component for open-world detection. 
To enhance the anti-catastrophic forgetting capability of these modules for better unseen categories detection ability, we propose a strong baseline {\method} for {\dataset} benchmark, a highly scalable open-world continual object detection method via \emph{\textbf{M}emory} and \emph{\textbf{R}etrieval} mechanisms. Specifically, {\method} employs a scalable memory pool, which efficiently caches parameter triplets regrading new concepts and visual-language interactions from continual learning steps. And during inference, {\method} enables to adaptively retrieve optimal parameter triplet to detect objects in previously learned, newly adapted, or open-world scenarios. The memory and retrieval mechanisms ensures the flexibility, scalability, and performance of {\method}, thus preserving detection capabilities of old, new and unseen open-world categories.

Extensive experiments are conducted on our proposed {\dataset} between different continual object detection frameworks and {\method}.
As shown in Fig~\ref{fig:intro}(c), with only tiny activated additional parameters our {\method} largely surpasses GDINO on seen classes with only few-shot continual adaptations.
Moreover, owning to robust retrieval machanism, {\method} enables simultaneous promising performance between  unseen and seen classes. 

In summary, our contributions are shown as follows:
\begin{itemize}
    \item We present {\dataset}, a challenging yet practical benchmark to assess seen and unseen classes detection abilities of OW detectors under few-shot continual adaptations. 
    \item We propose {\method}, a \emph{strong, efficient, and scalable} OW continual detector via \emph{memory and retrieval mechanisms} with a highly scalable memory pool. 
    \item By \emph{only 0.1\% activated extra parameters}, {\method} effectively improves detection capabilities for continually seen categories under few-shot adaptation, meanwhile ensuring open-world detection ability without forgetting. 
\end{itemize}

\section{Related Work}
\subsection{Open-World Object Detection}
Open-world (OW) object detection~\cite{li2021grounded,wang2024ov,zhang2022glipv2,bansal2018zero,ma2022open} aims to develop optimal detectors capable of recognizing both seen and unseen categories in real-world scenarios by vast semantic-rich multi-modal data~\cite{objects365,wang2023v3det,kazemzadeh2014referitgame,yu2016modeling,changpinyo2021cc12m,yao2022detclip,radford2021learning}.
A crucial component in OW detector design is the visual-language (VL) interaction module~\cite{li2021grounded,liu2023grounding,wang2024ov}, which links visual features with text embeddings, influencing detection capabilities. 
OW detectors are broadly classified into matching-based detectors~\cite{yao2022detclip,yao2023detclipv2} which use pretrained text embeddings to identify localized objects, and fusion-based detectors~\cite{li2021grounded,zhang2022glipv2,liu2023grounding,wang2024ov} 
which incorporate attention modules~\cite{li2021grounded,liu2023grounding} or ranking gates~\cite{wang2024ov} to merge visual and language features for accurate classification. However, seldom studies~\cite{deng2024zero} explore catastrophic forgetting in OW detectors under continual adaptations. 
In contrast, {\dataset} investigates this issue in continual adaptations, and {\method} ensures promising abilities on both seen and unseen classes.%

\noindent\textbf{Few-shot object detection with OW detectors. }
Our work shares similarities with few-shot object detection. Pretrained OW detectors~\cite{wang2020frustratingly,liu2023grounding} can adapt rapidly to target domains using few-shot training samples~\cite{dong2024lpt++,huang2024learning,wang2020frustratingly} for better performance. However, this often results in poor generalization to unseen categories~\cite{deng2024zero}. In contrast, {\method} demonstrates robust performance on both seen and unseen categories during continual few-shot adaptations.

\subsection{Continual Object Detection}

Continual object detection (COD)~\cite{dong2023incremental,liu2023continual,zhang2024dynamic} aims to learn detectors that incorporate new classes while retaining knowledge of prior ones. Early methods like ILOD\cite{shmelkov2017incremental} use pseudo-label distillation to address catastrophic forgetting~\cite{rebuffi2017icarl,li2017learning}, with recent works improving architectures and training strategies~\cite{liu2023continual,zhang2024dynamic,feng2022overcoming,song2024non}. However, few studies~\cite{deng2024zero} focus on OW-COD. In contrast, our {\method} introduces a retrieval-based~\cite{wang2022learning,dong2023lpt} approach for continual few-shot adaptations with pretrained OW models~\cite{liu2023grounding,wang2024ov}, preventing forgetting and extending COD to practical scenarios.

\section{Open-World Continual Object Detection}\label{sec:data}
\subsection{Task Definition}\label{sec:data_task}
Building upon COD and OWOD, we formulate the open-world continual object detection task. 
Given an open-world (OW) object detector $f$ pretrained on a large-scale dataset $\mathbb{D}_{\text{pre}}$, as well as a sequence of training set $\{\mathbb{D}_{1}, \dots, \mathbb{D}_{T}\}$ with size of $T$, 
we aim to optimize $f$ with corresponding parameters $\theta_{f}$ by sequentially learning on each $\mathbb{D}_{i}$. 
Such that the optimized $f(\cdot;\theta_{f})$ enables to accurately detect both previously learned old classes $\mathbb{C}_{1}\cup \dots \cup\mathbb{C}_{T\text{-}1}$ and newly learned classes $\mathbb{C}_{T}$, where $\mathbb{C}_{i}$ represents the label set of corresponding $\mathbb{D}_{i}$.
Meanwhile, $f(\cdot;\theta_{f})$ should be generalized well on open-world evaluation dataset $\mathbb{D}^{\text{val}}_{\text{unseen}}$ with corresponding large-scale and diverse label space $\mathbb{C}_{\text{unseen}}$.
The goal of our proposed task is to motivate OW detectors to simultaneously {\textbf{preserve} learned classes, \textbf{adapt} to new classes, and \textbf{maintain} open-world capabilities}, which is critical for OW detectors to simultaneously adapt to varying new environment and keep generalization abilities.

\subsection{Benchmark Construction}\label{sec:data_construct}
After defining the task, we formulate corresponding {\dataset} dataset as the data source for continual learning and universal evaluation for old, new, and unseen classes. 
Generally, {\dataset} is collected from existing object detection datasets~\cite{gupta2019lvis,li2021grounded} and broadly divided to two groups, \emph{i.e.}, seen category data and unseen category data. 
For seen category data, we leverage 13 subsets (from ``Aerial'' to ``Vehicle'') from ODinW-13~\cite{li2021grounded}, and assign $\{\mathbb{D}_{1}, \dots, \mathbb{D}_{T}\}$ by ascending dictionary order of subsets. The label space among $\{\mathbb{D}_{1}, \dots, \mathbb{D}_{T}\}$ are usually non-overlapped and fit the requirements of our task. During training of each step $t$, only images from $\mathbb{D}_{t}$ are visible. Notably, to simulate practical fast adaptation scenarios and increase the challenge of the benchmark, a few-shot training setting is adopted in {\dataset}. This setting requires continual OW detectors to effectively mitigate the impact of both overfitting and catastrophic forgetting, thereby enabling robust detection abilities for old, new, and unseen categories.
And for the unseen category data, to better align with real-world deployment scenarios, LVIS~\cite{gupta2019lvis} \emph{minival} set with $\sim$5k validation images and 1,203 categories is leveraged to empirically assess detection performance for unseen classes. This subset is only used for evaluation. Leveraging the dataset's large-scale and highly diverse label space facilitates empirical analysis of anti-forgetting abilities for unseen classes under continual adaptation. Statistics of the {\method} training and evaluation data are shown in the suppl.

\begin{figure*}[t]
\begin{center}
\includegraphics[width=\textwidth]{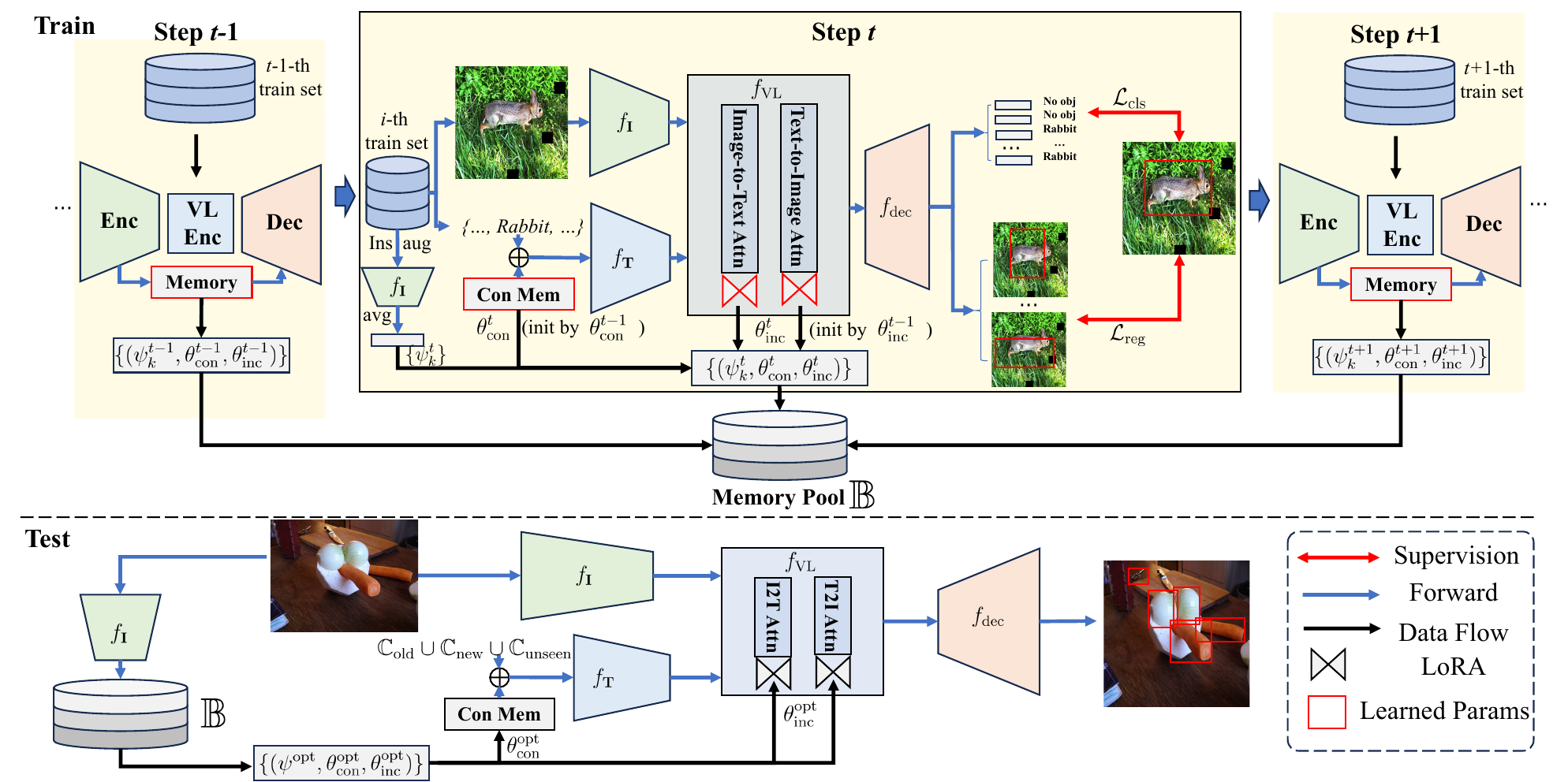}
\end{center}
\vspace{-1.5em}
\caption{
    Overview of our proposed {\method}. {\method} is based on a frozen pretrained open-world object detector with explicit visual-language interaction modules (\emph{e.g.}, Grounding DINO~\cite{liu2023grounding}). During each step $t$ of training, {\method} initializes concept memory $\theta^{t}_{\text{con}}$ and visual-language interaction memory $\theta^{t}_{\text{inc}}$ from corresponding parameters in the $t-1$ step, and optimizes both parameters by $t$-th training set. After training, $\theta^{t}_{\text{con}}$ and $\theta^{t}_{\text{inc}}$ are memorized into the memory pool $\mathbb{B}$. During open-world inference scenarios, {\method} uses the global embedding of input image $\mathbf{I}$ to retrieve the optimal parameters $(\psi^{\text{opt}}, \theta^{\text{opt}}_{\text{con}}, \theta^{\text{opt}}_{\text{inc}})$ and use these parameters for accurate predictions. 
}
\vspace{-1.5em}
\label{fig:pipeline}
\end{figure*}

\subsection{Metrics of {\dataset}}\label{sec:data_metric}

\noindent\textbf{Average Precision. }
Following the work on continual object detection~\cite{dong2023incremental,liu2023continual,deng2024zero,zhang2024dynamic} and open-world object detection~\cite{li2021grounded,wang2024ov,zhang2022glipv2,bansal2018zero,ma2022open}, the mean average precision (mAP) is reported for each subset to quantitatively assess the performance of learned open-world (OW) detectors under the continual learning paradigm. Specifically, per-subset average precision (AP) is provided to evaluate the detection performance of continual OW detectors after few-shot continual adaptations. Additionally, the mean AP for previously learned, newly seen, and unseen categories is reported to summarize the overall performance.

\noindent\textbf{Average rank. }Inspired by previous benchmarks~\cite{zhai2019large,li2021grounded}, {\dataset} also incorporates average rank as auxiliary metric to measure the relative performance of existing continual OW detectors. 
Specifically, {\dataset} first ranks all models within each subset. For the $K$ subsets of seen categories, let $R^i_j$ denote the rank of the $i$-th subset by the $j$-th detector. The average rank of seen categories $R^{\text{seen}}_j$ is then defined as:
\vspace{-0.5em}
\begin{equation}
\vspace{-0.5em}
    R^{\text{seen}}_{j} = \frac{\sum_{i=1}^{K}R^{i}_{j}}{K}.
\end{equation}
Similarly, we define the unseen classes average rank $R^{\text{unseen}}_{j}$ for $j$-th detector.
Finally, overall rank $R^{\text{avg}}_{j}$ is calculated by:
\vspace{-0.5em}
\begin{equation}
\vspace{-0.5em}
    R^{\text{avg}}_{j} = \sqrt{ \frac{{R^{\text{seen}}_{j}}^{2}+{R^{\text{unseen}}_{j}}^{2}}{2}}.
\end{equation}
The merit of this ranking lies in the fact that a detector can achieve a higher rank only if it performs well on both seen and unseen categories, thus emphasizing its ability to mitigate catastrophic forgetting for both old and new classes.

\subsection{Relation with Counterparts}\label{sec:data_relation}
\noindent\textbf{Comparison with COD. }
COD~\cite{dong2023incremental,liu2023continual,deng2024zero,zhang2024dynamic} typically split annotations from entire datasets (\emph{e.g.}, COCO~\cite{lin2014microsoft}) by dividing label sets into groups, which is not practical since novel categories often appear in unseen scenarios~\cite{wang2021wanderlust,continual_autodrive}, and seen images are usually fully labeled~\cite{objects365,lin2014microsoft,wang2023v3det} during annotation. {\method} avoids such irregular scenario.%
Besides, with the growing use of OW models in continual learning~\cite{yu2024boosting,wang2022learning}, {\dataset} emphasizes anti-forgetting capabilities for unseen categories. %

\noindent\textbf{Comparison with OWOD. }OWOD~\cite{li2021grounded,wang2024ov,zhang2022glipv2,bansal2018zero,ma2022open} can be seen as a zero-shot special case of our task. In contrast, {\dataset} simultaneously emphasizes the anti-forgetting capabilities among old, new, and unseen categories, which largely requires the generalization of OW detectors.

\noindent\textbf{Comparison with Deng \emph{et al.} }
Deng \emph{et al.}\cite{deng2024zero} conducted initial studies on open-world continual learning. However, their approach has two main drawbacks. First, the task-incremental evaluation is impractical for real-world applications and oversimplifies the challenge for continual open-world (OW) detectors. Second, their use of COCO~\cite{lin2014microsoft} for OW evaluation is limited, as it contains only 80 common classes that frequently recur across continual learning steps, thereby reducing the task's complexity.

\section{Proposed Method}\label{sec:method}
\subsection{Overview of {\method}}\label{sec:method_overview}
To accomplish the goal of {\dataset}, our core idea is \emph{first leveraging parameter-efficient modules to formulate ``memories'' in each step, then adaptively retrieving optimal memory for robust performance}.
Therefore, we propose {\method}, a strong baseline built upon the {\dataset} benchmark. {\method} utilizes a frozen open-world object detector with explicit visual-language interaction modules~\cite{li2021grounded,liu2023grounding} (\emph{e.g.}, Grounding DINO~\cite{liu2023grounding}) and incorporates memory and retrieval mechanisms for detection. The training and testing pipeline of {\method} is illustrated in Fig.~\ref{fig:pipeline}.
Specifically, during training of step $t$, given input image $\mathbf{I}$ and corresponding training label set $\mathbb{C}_{t}$, {\method} first concatenates class names in $\mathbb{C}_{t}$ by dot symbol, and formulate a unified class sentence $\mathbf{T}_{t}$. 
Then {\method} calculates dense image feature $\mathbf{F}_{\mathbf{I}}$ and text feature $\mathbf{F}_{\mathbf{T}}$ by image feature extractor $f_{\mathbf{I}}(\cdot; \theta_{\mathbf{I}})$ and text feature extractor $f_{\mathbf{T}}(\cdot; \theta_{\mathbf{T}}, \theta^{t}_{\text{con}})$ respectively, , where $\theta^{t}_{\text{con}}$ is the parameters of our proposed concept memory.
Next $\mathbf{F}_{\mathbf{I}}$ and $\mathbf{F}_{\mathbf{T}}$ are fed into the visual-language feature enhancer $f_{\mathbf{VL}}(\cdot; \theta_{\mathbf{VL}}, \theta^{t}_{\text{inc}})$ and obtain refined features $\mathbf{F}^{'}_{\mathbf{I}}$ and $\mathbf{F}^{'}_{\mathbf{T}}$, where $\theta^{t}_{\text{inc}}$ is the parameter of our proposed VL interaction memory. Finally, $\mathbf{F}^{'}_{\mathbf{I}}$ and $\mathbf{F}^{'}_{\mathbf{T}}$ are fed into the visual-language decoder $f_{\text{dec}}(\cdot; \theta_{\text{dec}})$ and obtain per-object detection results. Such results are supervised by corresponding ground-truth and used to optimize $\theta^{t}_{\text{con}}$ and $\theta^{t}_{\text{inc}}$.
And during inference, the input image $\mathbf{I}$ first extract the global embedding $\mathbf{g}_{\mathbf{I}}$ by image feature extractor $f_{\mathbf{I}}$. 
Then {\method} uses $\mathbf{g}_{\mathbf{I}}$ as query to retrieve the optimal memory triplets $\{(\psi^{\text{opt}}, \theta^{\text{opt}}_{\text{con}}, \theta^{\text{opt}}_{\text{inc}})\}$ from the memory pool $\mathbb{B}$ by threshold $\tau$. 
Finally, both input image $\mathbf{I}$ and class sentence $\mathbf{T}$ are fed into each OW detector $f(\cdot, \theta_{f}, \theta^{\text{opt}}_{\text{inc}})$ for initial detection results. These results are post-processed by Non-Maximum Suppression (NMS)~\cite{hosang2017learning} for final results. 
\subsection{Concept and Interaction Memory Mechanism}\label{sec:method_memory}
Inspired by parameter-efficient fine-tuning techniques in few-shot learning~\cite{dong2023lpt,jia2022vpt,zhou2022learning} and continual learning~\cite{yu2024boosting,dong2024consept,wang2022learning}, {\method} utilizes parameter-efficient modules as memory units (\emph{i.e.}, concept memory and visual-language (VL) interaction memory) for continually added classes to build optimal memories in corresponding learning steps. 

\noindent\textbf{Concept memory. }To make $f_{\mathbf{T}}(\cdot;\theta_{\mathbf{T}})$ adapt to continually added classes with negligible extra parameters, we introduce a learnable prompt $\theta_{\text{con}}$ into $f_{\mathbf{T}}$. During the $t$-th training step, with given class sentence $\mathbf{T}$, {\method} first convert $\mathbf{T}$ to initial text embedding $\mathbf{E}$ via embedding layer, and then concatenates both $\mathbf{E}$ and $\theta^{t}_{\text{con}}$,  finally the concatenated sequences are fed into transformer blocks in $f_{\mathbf{T}}$ and obtain the final text embedding $\mathbf{F}_{\mathbf{T}}$.

\begin{figure}[t]
  \begin{center}
  \includegraphics[width=0.4\textwidth]{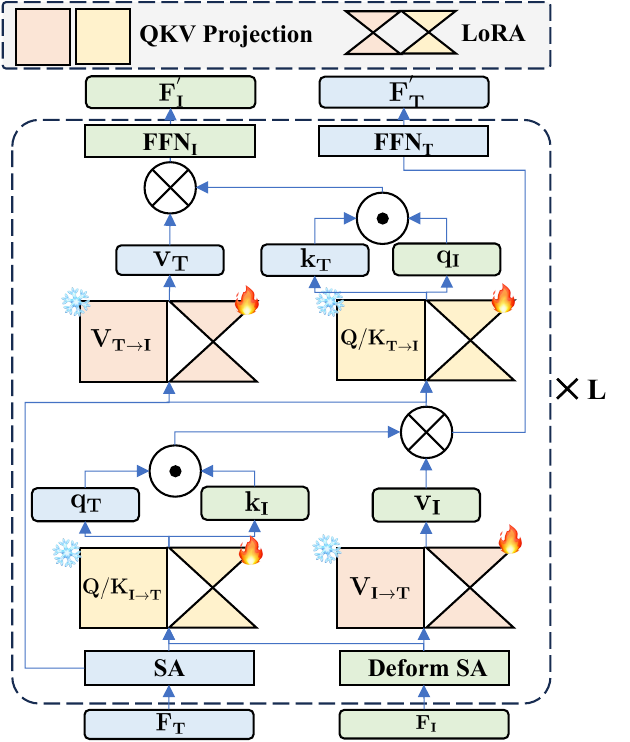}
  \end{center}
  \vspace{-1.5em}
  \caption{
      Overview of the proposed visual-language interaction memory. Specifically, {\method} adopts LoRA~\cite{hu2022lora} modules as $\theta_{\text{inc}}$ in Q/K/V projections of VL feature enhancer $f_{\mathbf{VL}}$.
  }
  \vspace{-1.5em}
  \label{fig:vli-mem}
  \end{figure}

\noindent\textbf{VL interaction memory. }Inspired by explicit VL interaction modules~\cite{li2021grounded,liu2023grounding}, we conclude that enhancing VL interaction of each step on these modules can lead to better continual OW detectors. To retrieve the optimal memory from memory pool for mitigating catastrophic forgetting, we propose VL interaction memory and leverage LoRA~\cite{hu2022lora} as corresponding memory, as shown in Fig.~\ref{fig:vli-mem}. 
In each $j$-th layer of $f_{\mathbf{VL}}$, given $\mathbf{F}_{\mathbf{I}}$ and $\mathbf{F}_{\mathbf{T}}$, {\method} uses deformable self-attention~\cite{zhu2021deformable} and vanilla self-attention~\cite{vaswani2017attention} to refine image and text features respectively, thereby obtaining $\hat{\mathbf{F}}_{\mathbf{I}}$ and  $\hat{\mathbf{F}}_{\mathbf{T}}$.
Then {\method} calculates aggregated text feature $\tilde{\mathbf{F}}_{\mathbf{T}}$ by:
\vspace{-0.2em}
\begin{equation}
\vspace{-0.3em}
   \begin{split}
   \tilde{\mathbf{F}}_{\mathbf{T}} &= \text{Attn}(\mathbf{q}_{\mathbf{T}}, \mathbf{k}_{\mathbf{I}}, \mathbf{v}_{\mathbf{I}})\text{, where} \\
   \mathbf{q}_{\mathbf{T}} &= (\mathbf{Q}_{\mathbf{I}\rightarrow\mathbf{T}}+\mathbf{B}^{\mathbf{q}}_{\mathbf{I}\rightarrow\mathbf{T}}\mathbf{A}^{\mathbf{q}}_{\mathbf{I}\rightarrow\mathbf{T}})\hat{\mathbf{F}}_{\mathbf{T}} \\
   \mathbf{k}_{\mathbf{I}} &= (\mathbf{K}_{\mathbf{I}\rightarrow\mathbf{T}}+\mathbf{B}^{\mathbf{k}}_{\mathbf{I}\rightarrow\mathbf{T}}\mathbf{A}^{\mathbf{k}}_{\mathbf{I}\rightarrow\mathbf{T}})\hat{\mathbf{F}}_{\mathbf{I}} \\
   \mathbf{v}_{\mathbf{I}} &= (\mathbf{V}_{\mathbf{I}\rightarrow\mathbf{T}}+\mathbf{B}^{\mathbf{v}}_{\mathbf{I}\rightarrow\mathbf{T}}\mathbf{A}^{\mathbf{v}}_{\mathbf{I}\rightarrow\mathbf{T}})\hat{\mathbf{F}}_{\mathbf{I}} \\
   \end{split}
\end{equation}
where ``Attn'' means cross-attention~\cite{vaswani2017attention}, $\mathbf{A}$ and $\mathbf{B}$ represent LoRA down- and up-projection layers. Note that $\mathbf{Q}$ and $\mathbf{K}$ with corresponding LoRA share the same parameters, and only $\mathbf{A}$ and $\mathbf{B}$ are optimized during training. 
Next, {\method} calculates the aggregated image feature $\tilde{\mathbf{F}}_{\mathbf{I}}$ by:
\vspace{-0.2em}
\begin{equation}
   \vspace{-0.3em}
   \begin{split}
   \tilde{\mathbf{F}}_{\mathbf{I}} &= \text{Attn}(\mathbf{q}_{\mathbf{I}}, \mathbf{k}_{\mathbf{T}}, \mathbf{v}_{\mathbf{T}})\text{, where} \\
   \mathbf{q}_{\mathbf{I}} &= (\mathbf{Q}_{\mathbf{T}\rightarrow\mathbf{I}}+\mathbf{B}^{\mathbf{q}}_{\mathbf{T}\rightarrow\mathbf{I}}\mathbf{A}^{\mathbf{q}}_{\mathbf{T}\rightarrow\mathbf{I}})\hat{\mathbf{F}}_{\mathbf{I}} \\
   \mathbf{k}_{\mathbf{T}} &= (\mathbf{K}_{\mathbf{T}\rightarrow\mathbf{I}}+\mathbf{B}^{\mathbf{k}}_{\mathbf{T}\rightarrow\mathbf{I}}\mathbf{A}^{\mathbf{k}}_{\mathbf{T}\rightarrow\mathbf{I}})\tilde{\mathbf{F}}_{\mathbf{T}} \\
   \mathbf{v}_{\mathbf{T}} &= (\mathbf{V}_{\mathbf{T}\rightarrow\mathbf{I}}+\mathbf{B}^{\mathbf{v}}_{\mathbf{T}\rightarrow\mathbf{I}}\mathbf{A}^{\mathbf{v}}_{\mathbf{T}\rightarrow\mathbf{I}})\tilde{\mathbf{F}}_{\mathbf{T}} \\
   \end{split}
\end{equation}
Finally, $\tilde{\mathbf{F}}_{\mathbf{I}}$ and $\tilde{\mathbf{F}}_{\mathbf{T}}$ are refined by corresponding feed-forward networks. After $L$-layer aggregation, one can obtain the final $\mathbf{F}^{'}_{\mathbf{I}}$ and $\mathbf{F}^{'}_{\mathbf{T}}$ for object detection. And the learned $\mathbf{A}$ and $\mathbf{B}$ in all layers formulate $\theta^{t}_{\text{inc}}$ in the $t$-th step.

\subsection{Memory Retrieval Mechanism}\label{sec:method_retrieval}
Both kind of memories can effectively incorporate knowledge regarding each step.
Nevertheless, such memories still faces catastrophic forgetting in both unseen and specific previously learned scenarios. 
To mitigate this issue, an intuitive idea is explicitly memorizing all previously learned memories, and adaptively retrieving best matched modules during inference. Such method is also aligned with human memory~\cite{baddeley1997human,cowan2012working}.
Therefore, we propose the retrieval mechanism. Specifically, {\method} introduces a memory pool $\mathbb{B}$ to store all previously learned memories. For the $t$-th step, with given $n$-shot training images, {\method} first augment images by instance cropping and obtain totally $N$ images for each class $k$ in $\mathbb{C}_{t}$. Then {\method} utilizes $f(\cdot; \theta_{I})$ to calculate each averaged global embedding $\psi^{t}_{k}$ averaging operation. 
And finally, {\method} formulates the triplet of $t$-th step by $\{(\psi^{t}_{k}, \theta^{t}_{\text{con}}, \theta^{t}_{\text{inc}})\}$, where $\theta^{t}_{\text{con}}$ and $\theta^{t}_{\text{inc}}$ are optimized concept and VL interaction memory from $t$-th step. 
And during inference, given input image $\mathbf{I}$, {\method} first extract the global embedding $\mathbf{g}_{\mathbf{I}}$, then we find the indices $\{\hat{t}\}$ of memory triplets by threshold $\tau$, and ensure that $\{\hat{t}\} = \{t|\langle \mathbf{g}_{\mathbf{I}},\psi^{t} \rangle>\tau\}$. Finally, we retrieve the optimal memories 
$\{(\psi^{\text{opt}}, \theta^{\text{opt}}_{\text{con}}, \theta^{\text{opt}}_{\text{inc}})\}$ by:
\vspace{-0.5em}
\begin{equation}\label{eqn:merge_pseudo_gt}
\vspace{-0.5em}
    (\psi^{\text{opt}}, \theta^{\text{opt}}_{\text{con}}, \theta^{\text{opt}}_{\text{inc}}) = \left\{
    \begin{aligned}
        & \{(\psi^{\hat{t}}, \theta^{\hat{t}}_{\text{con}}, \theta^{\hat{t}}_{\text{inc}})\} & \langle \mathbf{g}_{\mathbf{I}},\psi^{\hat{t}} \rangle \geqslant \tau \\
        & \phi & \{\hat{t}\} = \phi
    \end{aligned}\right .
\end{equation}
Such design ensures that, when unseen objects occurs, {\method} enables to use the vanilla pretrained OW detector to detect objects in the wild, thus preserving the detection abilities of open-world unseen categories. 
\subsection{Training of {\method}}\label{sec:method_training}

During training, parameters of pretrained OW detector are frozen to preserve robust feature representation~\cite{wang2022learning,jia2022vpt,dong2023lpt}, while only concept and VL interaction memories are optimized. Specifically, to maintain a consistent text embedding distribution from frozen $f_{\mathbf{T}}$ for stable training, memory training is divided into two stages. In the first stage, {\method} freezes VL interaction memory and optimizes concept memory to adapt to new classes. In the second stage, the updated concept memory is frozen, and interaction memory is optimized to refine visual-language relationships. Notably, joint training of both memory types can achieve similar performance, as discussed in Sec.~\ref{sec:exp_ablation}.

\begin{table}[t]
  \centering
      \caption{Comparison between {\method} and counterparts. {\method} simultaneously merits from flexibility, scalability, efficiency, thus achieving better anti-forgetting ability.}
      \vspace{-1em}
      \label{table:merit}
      \small
      \setlength{\tabcolsep}{2pt} %
     \renewcommand{\arraystretch}{4.0}%
   { \fontsize{8.3}{3}\selectfont{
      \begin{tabular}{c|ccc|c}
      \bottomrule
      \textbf{Method}&\textbf{Flexibility}&\textbf{Scalability} & \textbf{Efficiency}&\textbf{OW Anti-forget}
      \\ \hline
       CoOp~\cite{zhou2022learning} & \XSolidBrush & \XSolidBrush  &\Checkmark & \XSolidBrush  \\
       L2P~\cite{wang2022learning}& \Checkmark  &\XSolidBrush & \Checkmark & \XSolidBrush  \\
       ZiRa~\cite{deng2024zero} & \Checkmark & \XSolidBrush  & \Checkmark & \XSolidBrush \\
       CL-DETR~\cite{liu2023continual} & \XSolidBrush  & \XSolidBrush  & \XSolidBrush  & \XSolidBrush \\
       {\method}& \Checkmark & \Checkmark & \Checkmark & \Checkmark \\
      \toprule
      \end{tabular}
      }}
      \vspace{-2em}
  \end{table}

\noindent\textbf{Training Objectives. }
Unlike previous works~\cite{liu2023continual,dong2023incremental}, {\method} does not use additional losses specifically designed for continual learning. For bounding box regression, {\method} minimizes L1 loss and GIoU loss~\cite{Rezatofighi_2018_CVPR} at each training step. For object classification, focal loss~\cite{Lin_2020} is employed to enhance recognition performance.

\begin{table*}[t]
  \centering
      \caption{
      Comparison of diverse open-world continual learning frameworks . We keep the pretrained models of all frameworks are the same Grounding DINO with Swin-T. Best results are \textbf{bolded} and second best results are \underline{underlined}. Compared to zero-shot GDINO, all the baselines face severe catastrophic forgetting on either seen classes or open-world unseen classes. In contrast, {\method} expresses promising anti-forgetting capabilities on both seen and unseen classes, and surpasses all the counterparts in terms of detection abilities. 
      }
      \label{table:comp_sota}
      \vspace{-1em}
      \small
      \setlength{\tabcolsep}{5.0pt} %
     \renewcommand{\arraystretch}{4.0}%
   { \fontsize{8.3}{3}\selectfont{
      \begin{tabular}{cl|ccccccccccccc|cc|c}
      \bottomrule
      \textbf{Shot} & \textbf{Method} & \textbf{Ae} & \textbf{Aq} & \textbf{Co} & \textbf{Eg} & \textbf{Mu} & \textbf{Pa} & \textbf{VOC} & \textbf{Pi} & \textbf{Po} & \textbf{Ra} & \textbf{Sh} & \textbf{Th} & \textbf{Ve} & \textbf{Seen} & \textbf{Unseen} & $R^{\text{avg}}$ \\
      \hline
      0&ZS GDINO& 15.4 & 18.4 & 69.0 & 57.7 & 14.8 & 65.3 & 50.2 & 53.8 & 16.1 & 0 & 22.8 & 42.6 & 58.8 & 37.3 & 20.7 & - \\
      
      \hline
      \multirow{5}{*}{1}&CoOp&\underline{14.6}&15.7&\underline{73.6}&0&\underline{58.1}&63.3&\textbf{54.2}&\textbf{49.7}&9.6&\underline{27.6}&\textbf{31.9}&\underline{57.5}&\textbf{59.6}&\underline{39.6}&\underline{20.5}&\underline{2.1} \\
      &Adapter& 13.7 &\underline{17.6}&66.2&\underline{55.8}&44.9&67.4&48.4&41.4&12.3&0&20.8&45.7&55.5&37.7&19.7&3.3 \\
      &L2P&11.5&16.9&71.0&2.5&45.2&53.6&45.3&48.4&3.0&25.2&22.2&37.8&\underline{59.4}&34.0&18.7&3.9 \\
      &ZiRa&11.4&12.4&0&9.8&\textbf{66.1}&\underline{68.6}&39.1&32.7&2.0&\textbf{48.7}&23.4&57.1&50.4&32.5&6.9&4.4 \\
      
      &{\method}& \textbf{20.0} & \textbf{20.8} & \textbf{76.2} & \textbf{59.6} & {56.3} &\textbf{69.0} & \underline{51.5} & \underline{49.3} & \textbf{19.7} & 26.3 & \underline{28.4} & \textbf{71.1} & 58.9 & \textbf{46.7} & \textbf{20.6} & \textbf{1.3} \\

      \hline
      \multirow{5}{*}{3}&CoOp&10.5 &3.7&0&0&30.6&59.9&8.3&39.5&1.4&12.8&14.3&32.1&47.1&20.1&19.4&3.6 \\
      
      &Adapter&\underline{14.8} &\underline{18.4}&68.9&\underline{55.7}&\underline{47.4}&\underline{67.4}&49.2&39.9&\underline{12.5}&0&22.6&51.4&58.0&\underline{39.0}&\underline{19.4}&\underline{2.4} \\
      
      &L2P&13.2 &15.8&\underline{74.3}&15.3&36.5&66.4&\underline{50.0}&43.8&2.1&8.5&18.7&46.6&\textbf{63.1}&35.0&18.8&3.6 \\
      
      &ZiRa&13.4&14.3&0&2.6&50.0&60.7&47.1&\underline{51.2}&7.3&\textbf{52.9}&\textbf{33.4}&\underline{56.6}&42.4&33.2&7.3&4.2 \\
      
      &{\method}& \textbf{28.6} & \textbf{26.2} & \textbf{76.5} & \textbf{67.9} & \textbf{73.1} & \textbf{68.2} & \textbf{50.9} & \textbf{58.8} & \textbf{22.2} & \underline{30.8} & \underline{25.6} & \textbf{70.6} & \underline{59.5} & \textbf{50.7} & \textbf{20.6} &\textbf{1.1}  \\

      \hline
      \multirow{5}{*}{5}&CoOp& 9.8 & 12.5 &42.2&0&\underline{56.2}&55.1&28.2&22.6&3.9&30.0&\underline{25.7}&28.0&61.3&28.9&19.1&3.5 \\
      &Adapter&\underline{14.4} &\underline{18.7}&\underline{69.3}&\underline{56.8}&47.1&\underline{67.3}&\underline{49.7}&41.7&\underline{13.1}&0&23.1&49.0&57.6&\underline{39.1}&\underline{20.2}&\underline{2.5} \\
      &L2P& 11.3 &14.3&50.3&0&42.8&59.9&35.8&52.1&3.9&27.2&23.8&35.3&\textbf{64.4}&32.4&17.4&3.7 \\
      &ZiRa&12.5 &9.2&44.5&0&38.0&59.7&47.8&\underline{55.9}&4.4&\underline{34.9}&\textbf{30.2}&\textbf{56.6}&\underline{62.9}&35.1&5.8&4.1 \\
      &{\method}& \textbf{28.7} & \textbf{26.3} & \textbf{80.6} & \textbf{69.4} & \textbf{60.1} & \textbf{74.0} & \textbf{50.8} & \textbf{63.8} & \textbf{25.4} & \textbf{43.3} & 23.6 & \underline{52.7} & 62.0 & \textbf{50.8}  & \textbf{20.6} & \textbf{1.3} \\
      
      \hline
      \multirow{5}{*}{10}&CoOp& 11.9&16.3&56.1&0.4&57.6&59.5&44.0&45.3&4.9&19.1&23.6&46.6&\underline{62.7}&34.5&17.0&3.7 \\
      &Adapter& \underline{16.8}&\underline{18.1}&71.8&\underline{54.7}&37.0&66.1&\textbf{50.5}&35.4&11.7&0&25.1&40.0&58.2&37.3&\underline{20.4}&\underline{2.7} \\
      &L2P& 9.1&12.9&22.8&0.9&41.3&50.3&30.3&41.0&\underline{11.7}&9.0&19.0&37.8&61.8&26.8&17.8&3.9 \\
      &ZiRa& 10.7&6.5&\underline{74.9}&0&\underline{66.0}&\textbf{69.8}&46.0&\underline{49.7}&6.0&\textbf{40.3}&\textbf{32.4}&\underline{59.6}&\textbf{63.2}&\underline{40.4}&6.9&4.0 \\
       &{\method}& \textbf{30.5} & \textbf{26.1} & \textbf{79.4} & \textbf{65.3} & \textbf{75.3} & \underline{67.1} & \underline{48.3} & \textbf{65.0} & \textbf{30.4} & \underline{27.4} & \underline{25.7} & \textbf{74.8} & 59.7 & \textbf{51.9} & \textbf{20.7} & \textbf{1.3} \\
      \toprule
      \end{tabular}
      }}
      \vspace{-1em}
  \end{table*}
  
\subsection{Relation with Counterparts and Merits}\label{sec:method_merits}

As shown in Table~\ref{table:merit}, {\method} excels in three aspects. For \textbf{flexibility}, it outperforms CoOp and CL-DETR with flexible memory retrieval through activated parameter selection. For \textbf{scalability}, {\method} surpasses L2P~\cite{wang2022learning} with a scalable memory pool that preserves and integrates knowledge. Lastly, for \textbf{efficiency}, {\method} leverages parameter-efficient fine-tuning, outperforming traditional full fine-tuning methods~\cite{liu2023continual,zhang2024dynamic,Chen2019ANK}. These strengths ensure strong performance on old, new, and unseen classes.

\section{Experiments}\label{sec:exp}
We compare {\method} with zero-shot GDINO~\cite{liu2023grounding}, CoOp~\cite{zhou2022learning}, L2P~\cite{wang2022learning}, Adapter~\cite{adapter}, and ZiRa~\cite{deng2024zero}. All methods are designed for continual or fast adaptation. 
\subsection{Implementation Details}\label{sec:exp_implementation}
We employ the Swin-T~\cite{Liu_2021_ICCV} Grounding DINO~\cite{liu2023grounding} as the pretrained OW detector for both {\method} and the counterparts. For continual training on {\dataset}, we optimize OW detectors following the ascending dictionary order of subsets and evaluate the trained detectors on old, new, and unseen categories from corresponding subsets without any test-time augmentation. For {\method}, we set a default prompt length of 10 and a LoRA~\cite{hu2022lora} bottleneck dimension of 8. We use AdamW~\cite{adamw} with cosine learning rate scheduler~\cite{loshchilov2017sgdr,goyal2017accurate} to optimize {\method} with weight decay of 1e-2 and batch size of 1 per GPU. The initial learning rate candidates are \{1e-1, 4e-2, 1e-2, 1e-3, 1e-4\}, and training epochs range from \{1$\sim$10\}. We perform grid search~\cite{lavalle2004relationship} to find optimal hyper-parameters for each step. $\tau$ is set to 0.89 by default. Baseline methods are constructed and optimized using their default hyper-parameters. Due to the absence of an LVIS evaluation toolkit in the original GDINO implementation, we implement corresponding toolkit to fairly assess old, new, and unseen classes across all methods. %

\begin{table*}[t]
  \centering
      \caption{
      Comparison between Grounding DINO after COCO fully fine-tuning and that with {\method} (10-shot), where \textcolor{red}{red} means subsets with performance drop after fine-tuning. Compared to zero-shot Grounding DINO and GDINO (COCO-ft), {\method} enables to mitigate forgotten classes by few-shot continual adaptations, meanwhile preserving promising detection abilities on unseen classes. 
      }
      \label{table:mitigate_forgotten}
      \vspace{-1em}
      \small
      \setlength{\tabcolsep}{5.0pt} %
     \renewcommand{\arraystretch}{4.0}%
   { \fontsize{8.3}{3}\selectfont{
      \begin{tabular}{l|c|ccccccccccccc|cc}
      \bottomrule
      \textbf{Method} & COCO & \textcolor{red}{\textbf{Ae}} & \textbf{Aq} & \textcolor{red}{\textbf{Co}} & \textcolor{red}{\textbf{Eg}} & \textbf{Mu} & \textbf{Pa} & \textbf{VOC} & \textcolor{red}{\textbf{Pi}} & \textcolor{red}{\textbf{Po}} & \textbf{Ra} & \textbf{Sh} & \textbf{Th} & \textcolor{red}{\textbf{Ve}} & \textbf{Seen} & \textbf{Unseen}  \\
      \hline
      ZS GDINO & 48.4 & 15.4 & 18.4 & 69.0 & 57.7 & 14.8 & 65.3 & 50.2 & 53.8 & 16.1 & 0.0 & 22.8 & 42.6 & 58.8 & 37.3 & 20.7  \\
      
      \hline
      GDINO (COCO-ft) & \textbf{57.3} &12.5&20.7&64.1&34.8&43.6&\textbf{66.4}&52.0&48.5&7.2&0.0&35.0&46.0&54.2&37.3&23.6 \\
        +{\method}& 57.2 &  \textbf{20.7} & \textbf{29.2} & \textbf{82.1} & \textbf{64.8} & \textbf{83.3} & {66.2} & \textbf{62.2} & \textbf{65.1} & \textbf{16.2} & \textbf{50.2} & \textbf{40.2} & \textbf{69.8} & \textbf{59.6} & \textbf{54.5} & \textbf{23.6}  \\

      \toprule
      \end{tabular}
      }}
      \vspace{-1.5em}
  \end{table*}

\begin{figure*}[t]
\begin{center}
\includegraphics[width=0.9\textwidth]{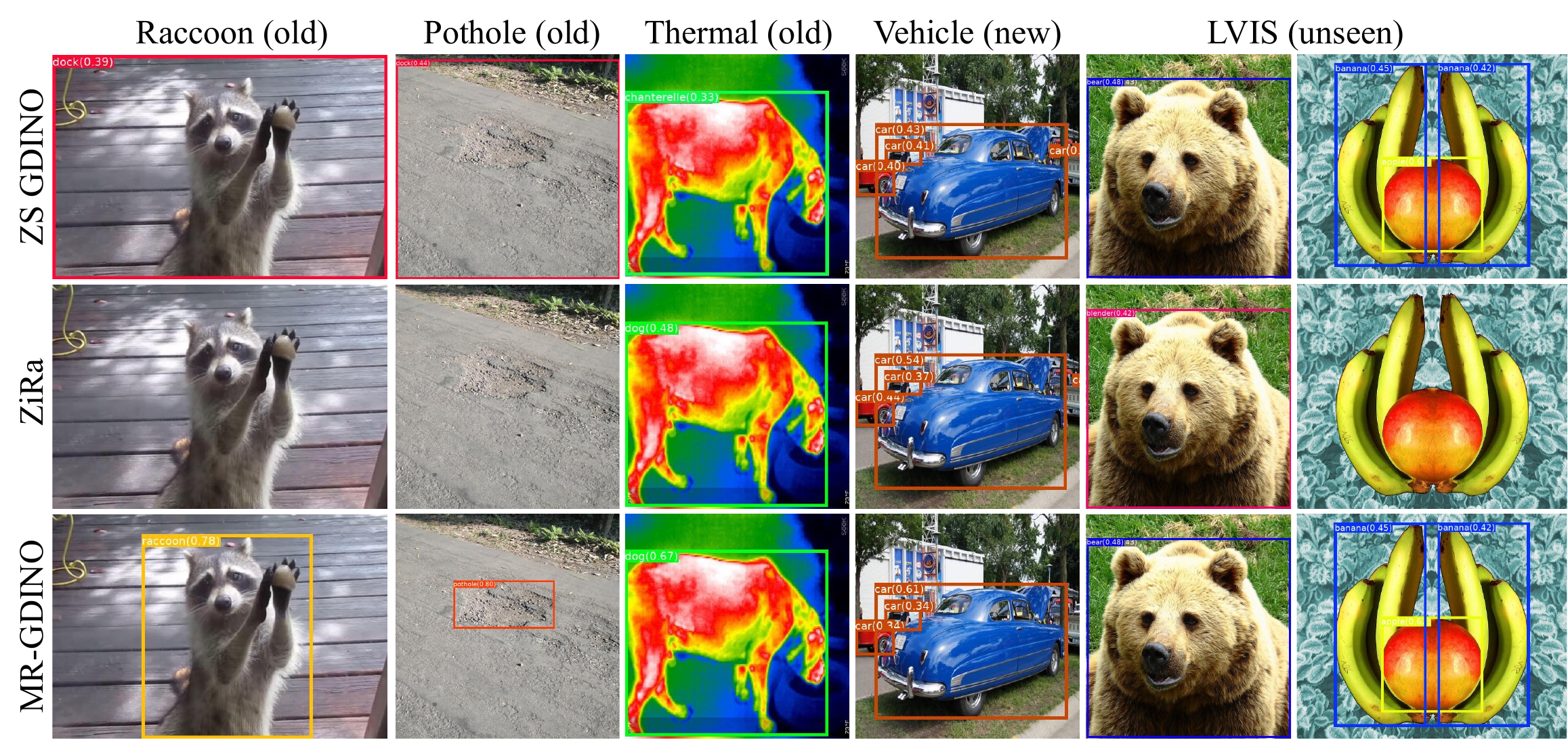}
\end{center}
\vspace{-1.5em}
\caption{
    Qualitative results of zero-shot Grounding DINO~\cite{liu2023grounding}, ZiRa~\cite{deng2024zero}, and {\method}. Compared to ZS GDINO and state-of-the-art ZiRa, {\method} can generate more accurate bounding boxes with higher confidence on both seen and unseen classes. 
}
\vspace{-1.5em}
\label{fig:vis}
\end{figure*}

\subsection{Comparison with State-of-The-Arts}\label{sec:exp_comp}

Table~\ref{table:comp_sota} presents the comparison between {\method} and all the counterparts under continual adaptations with different shots.
Among all the counterparts, only ZiRa~\cite{deng2024zero} surpasses ZS GDINO by 3.1 on $\text{AP}^{\text{seen}}$ after 10-shot continual adaptations, while other methods fail to outperform GDINO. For unseen classes, only the Adapter~\cite{adapter} based continual OW detector achieves comparable albeit lower mAP, with all other methods suffering significant catastrophic forgetting. These findings strongly support our perspective and highlight the importance of OW-COD.
In contrast, {\method} under 10-shot training achieves 51.9 $\text{AP}^{\text{seen}}$ and 20.7 $\text{AP}^{\text{unseen}}$.
Moreover, even in 1-shot continual learning settings, {\method} still achieves 46.7 seen mAP and only suffers 0.1 drop in terms of unseen mAP, and still largely surpasses all the counterparts on both metrics.
Such promising results demonstrate that {\method} can largely improves the detection performance on old and new classes, meanwhile maintaining robust detection abilities for unseen categories. We also investigate corresponding forgetting rate in each training step, which is listed in the suppl.
Furthermore, though ZiRa and Adapter show improved anti-forgetting abilities for seen and unseen categories, respectively, their average rankings remain affected by the imbalanced performance between seen and unseen classes. In contrast, {\method} achieves rank 1.3 in terms of $R^{\text{avg}}$ on the leaderboard, underscoring its balanced and superior performance across old, new, and unseen classes.

\noindent\textbf{Qualitative Results. }%
Besides, we present qualitative results among ZS GDINO~\cite{liu2023grounding}, ZiRa~\cite{deng2024zero}, and {\method}, as shown in Fig.~\ref{fig:vis}. Notably, {\method} produces accurate bounding boxes with higher confidence for both old and new classes. Moreover, {\method} outperforms ZiRa in generating accurate bounding boxes for unseen classes. These results further confirm the effectiveness of {\method}.
More qualitative results are shown in the suppl.

\subsection{{\method} Can Mitigate Forgotten Classes}\label{sec:exp_cocoft}
Based on the promising anti-forgetting capabilities in both seen and unseen classes, one can leverage {\method} to mitigate ``forgotten'' classes from fine-tuning. Specifically, we fully fine-tune GDINO~\cite{liu2023grounding} on COCO~\cite{lin2014microsoft}, and corresponding evaluation results are shown in Table~\ref{table:mitigate_forgotten}. Though detection performance on COCO increases to 57.3 mAP, detection performance on 6 out of 13 subsets has dropped, which can be seen as forgotten unseen classes. By adopting {\method} onto GDINO (COCO-ft), detection performance on above subsets has increased and achieves 54.5 $\text{AP}^{\text{seen}}$. Meanwhile, since COCO~\cite{lin2014microsoft} and LVIS~\cite{gupta2019lvis} has large overlap in image domain, the $\text{AP}^{\text{unseen}}$ of GDINO (COCO-ft) has increased to 23.6 due to fully fine-tuning. Compared to GDINO (COCO-ft), that with MR-GDINO preserves the same $\text{AP}^{\text{unseen}}$. Above results further verify the effectiveness of {\method} in mitigating forgetting.
\subsection{Empirical Analysis}\label{sec:exp_ablation}

\subsubsection{Ablation Study of Each Component}
We first conduct ablation study of each component with 10-shot continual learning. Table~\ref{table:ablation} demonstrates the evaluation results of each method. 
After adopting $\theta_{\text{con}}$, $\text{AP}^{\text{old}}$ and $\theta_{\text{con}}$, $\text{AP}^{\text{unseen}}$ largely drops to 32.2 and 17.0 respectively, but $\text{AP}^{\text{new}}$ largely increases to 62.1. Similarly, when further adopting $\theta_{\text{inc}}$ into {\method}, corresponding $\text{AP}^{\text{new}}$ increases to 63.1. 
Above optimized memories provide strong and robust learned parameters on each subset and will benefit retrieval mechanism. After adopting retrieval mechanism, both $\text{AP}^{\text{old}}$ and $\text{AP}^{\text{unseen}}$ significantly increase to 51.3 and 20.7 respectively, which indicates that such mechanism can effectively retrieve optimal $\theta_{\text{con}}$ and $\theta_{\text{inc}}$ for given inputs to achieve better detection abilities. And if the input images come from unseen categories, {\method} can still execute correct action and use ZS GDINO for inference. 
These findings verify the effectiveness of memory and retrieval mechanisms in {\dataset}, and reveal potential directions towards better continual OW detectors.

\begin{table}[t]
  \centering
      \caption{Ablation study of key components in {\method}. }
      \vspace{-1em}
      \label{table:ablation}
      \small
      \setlength{\tabcolsep}{4pt} %
     \renewcommand{\arraystretch}{4.0}%
   { \fontsize{8.3}{3}\selectfont{
      \begin{tabular}{cccc|cccc}
      \bottomrule
      GDINO&$\theta_{\text{con}}$&$\theta_{\text{inc}}$ & retrieval&$\text{AP}^{\text{old}}$&$\text{AP}^{\text{new}}$&$\text{AP}^{\text{seen}}$&$\text{AP}^{\text{unseen}}$\\
       \hline
       \Checkmark & \XSolidBrush & \XSolidBrush  &\XSolidBrush & 35.5 & 58.8 & 37.3 & 20.7  \\
       \Checkmark& \Checkmark  &\XSolidBrush & \XSolidBrush & 32.2 & 62.1 & 34.5 & 17.0 \\
       \Checkmark & \Checkmark & \XSolidBrush  & \Checkmark & 44.4 & 56.8 & 45.4 & 20.7 \\
       \Checkmark & \Checkmark & \Checkmark  & \XSolidBrush & 30.9 & 63.1 & 33.3 & 10.7 \\
     \Checkmark & \Checkmark & \Checkmark & \Checkmark & 51.3 & 59.7 & 51.9 & 20.7 \\
      \toprule
      \end{tabular}
      }}
      \vspace{-1.5em}
  \end{table}

\begin{table}[t]
  \centering
      \caption{Number of $\theta_{\text{inc}}$-inserted layers analysis. 
      }
      \vspace{-1em}
      \label{table:ablation_L}
      \small
      \setlength{\tabcolsep}{2.5pt} %
     \renewcommand{\arraystretch}{4.0}%
   { \fontsize{8.3}{3}\selectfont{
      \begin{tabular}{lcc|cccc}
      \bottomrule
      Method&Layers&Added Params&$\text{AP}^{\text{old}}$&$\text{AP}^{\text{new}}$&$\text{AP}^{\text{seen}}$&$\text{AP}^{\text{unseen}}$\\
       \hline
       {\method}&0& 12K  & 44.4 & 56.8 & 45.4 & 20.7 \\
       {\method}&1& 53K  & 47.3 & 57.5 & 48.1 & 20.7 \\
       {\method}&2& 94K  & 49.0 & 58.2 & 49.7& 20.7 \\
       {\method}&3& 135K  & 50.1 & 59.9 & 50.8 & 20.7 \\
      {\method}&6 (all)& 258K  & 51.3 & 59.7 & 51.9 & 20.7 \\
      \toprule
      \end{tabular}
      }}
      \vspace{-1.5em}
  \end{table}

\subsubsection{Effect of Inserted Layer Number for $\theta_{\text{inc}}$}
Next we investigate the effect from inserted layer number of $\theta_{\text{inc}}$, corresponding results are shown in Table~\ref{table:ablation_L}. By inserting $\theta_{\text{inc}}$ in more layers, $\text{AP}^{old}$ are gradually increased from 44.4 to 51.3, while maintaining the same $\text{AP}^{\text{unseen}}$. These results show that inserting $\theta_{\text{inc}}$ to more VL interaction layers lead to better performance with negligible parameters.

\begin{table}[t]
  \centering
      \caption{Comparison between joint and decoupled training. 
      }
      \vspace{-1em}
      \label{table:joint_train}
      \small
      \setlength{\tabcolsep}{5pt} %
     \renewcommand{\arraystretch}{4.0}%
   { \fontsize{8.3}{3}\selectfont{
      \begin{tabular}{cc|cccc}
      \bottomrule
      Joint&Decouple&$\text{AP}^{\text{old}}$&$\text{AP}^{\text{new}}$&$\text{AP}^{\text{seen}}$&$\text{AP}^{\text{unseen}}$\\
       \hline
       \Checkmark & - &  51.6 & 59.7 & 52.2 & 20.7  \\
      -& \Checkmark  & 51.3 & 59.7 & 51.9 & 20.7 \\
      \toprule
      \end{tabular}
      }}
      \vspace{-1.5em}
  \end{table}

\begin{table}[t]
  \centering
      \caption{Performance gap between {\method} and {\method} with oracle retrieval module.}
      \vspace{-1em}
      \label{table:ablation_retrieval}
      \small
      \setlength{\tabcolsep}{6pt} %
     \renewcommand{\arraystretch}{4.0}%
   { \fontsize{8.3}{3}\selectfont{
      \begin{tabular}{c|cccc}
      \bottomrule
      Retrieval&$\text{AP}^{\text{old}}$&$\text{AP}^{\text{new}}$&$\text{AP}^{\text{seen}}$&$\text{AP}^{\text{unseen}}$\\
       \hline
        
       {\method}  & 51.3 & 59.7 & 51.9 & 20.7 \\
       Oracle &  51.8 & 59.8 & 52.4 & 20.7  \\
      \toprule
      \end{tabular}
      }}
      \vspace{-2em}
  \end{table}

\subsubsection{Decoupled Training or Joint Training}
We also investigate whether {\method} supports joint training for $\theta^{t}_{\text{con}}$ and $\theta^{t}_{\text{inc}}$ at each training step $t$. Using the optimal training hyper-parameters identified from decoupled training, we simultaneously optimize $\theta^{t}_{\text{con}}$ and $\theta^{t}_{\text{inc}}$. The results, shown in Table~\ref{table:joint_train}, indicate that joint training achieves the same 59.7 $\text{AP}^{\text{new}}$ and $\text{AP}^{\text{unseen}}$, with slightly improved $\text{AP}^{\text{old}}$. These findings suggest that once optimal hyper-parameters are confirmed, joint optimization can halve the training time to improve efficiency. 

\subsubsection{Performance Gap with Oracle Retrieval}
Finally, we analyze the retrieval mechanism to assess the performance gap between {\method} and oracle counterparts. For the oracle retrieval, we assign $\theta^{\text{opt}}_{\text{con}}$ and $\theta^{\text{opt}}_{\text{inc}}$ using ground-truth labels and report detection results in Table~\ref{table:ablation_retrieval}. Compared to the oracle, {\method} shows a minor decrease of 0.5 in $\text{AP}^{\text{old}}$ and 0.1 in $\text{AP}^{\text{new}}$, while achieving similar performance in $\text{AP}^{\text{unseen}}$. These results confirm the effectiveness of {\method}’s retrieval mechanism. However, exploring more precise retrieval mechanisms remains valuable for future large-scale and practical applications. Further analysis is provided in the supplementary material.

\section{Conclusion}\label{sec:conclusion}
We propose open-world continual object detection, requiring detectors to generalize across old, new, and unseen categories. To evaluate OW detectors with existing continual learning methods, we propose {\dataset}, a benchmark encouraging OW detectors to {preserve} old classes, {adapt} to new ones, and {maintain} open-world detection abilities. To address catastrophic forgetting of unseen categories, we propose a strong baseline namely {\method}, a scalable open-world object detection framework utilizing {memory and retrieval} in a compact memory pool. Our results show that {\method} minimizes catastrophic forgetting with only 0.1\% additional parameters, achieving state-of-the-art performance on {\dataset}.

{
    \small
    \bibliographystyle{ieeenat_fullname}
    \bibliography{main}
}

\end{document}